\def\arxivpreprint{1}
\documentclass{article}

\usepackage[T1]{fontenc}
\usepackage{amsmath}
\usepackage{amssymb}
\usepackage{booktabs}
\usepackage{graphicx}
\usepackage{microtype}
\usepackage{multirow}
\usepackage{enumitem}
\usepackage[hypertexnames=false]{hyperref}
\usepackage{xurl}

\ifdefined\arxivpreprint
\usepackage[accepted]{mlsys2025}
\makeatletter
\renewcommand{\Notice@String}{Preprint.}
\makeatother
\else
\usepackage{mlsys2025}
\fi

\newcommand{\method}{Trie-Head}

\newcommand{\supplementref}[1]{%
\ifdefined\arxivpreprint Appendix~\ref{#1}\else Supplementary material\fi}

\mlsystitlerunning{wide-beam generative retrieval}

\begin{document}
\twocolumn[
\mlsystitle{Efficient Wide-Beam Generative Retrieval with Model-Native Trie Decoding and Shared-Context Attention}

\begin{mlsysauthorlist}
\mlsysauthor{Vedant Jhaveri}{linkedin}
\mlsysauthor{Arashpreet Singh Mor}{linkedin}
\mlsysauthor{Satyam Kumar}{linkedin}
\mlsysauthor{Alex Tsun}{linkedin}
\mlsysauthor{Qing Lan}{linkedin}
\mlsysauthor{Shaunak Mishra}{linkedin}
\mlsysauthor{Sriram Vasudevan}{linkedin}
\end{mlsysauthorlist}

\mlsysaffiliation{linkedin}{LinkedIn}
\mlsyscorrespondingauthor{Vedant Jhaveri}{vjhaveri@linkedin.com}
\mlsyskeywords{Generative Retrieval, Constrained Decoding, LLM Serving, GPU Systems}
\ifdefined\arxivpreprint
\hypersetup{pdfauthor={Vedant Jhaveri, Arashpreet Singh Mor, Satyam Kumar, Alex Tsun, Qing Lan, Shaunak Mishra, Sriram Vasudevan}}
\fi

\vskip 0.3in

\begin{abstract}
Generative retrieval (GR) predicts catalog entries as short semantic
identifiers (SIDs). Trie-constrained beam search keeps these identifiers
valid, but wide beams expose two serving costs: vocabulary-wide score
tensors that are mostly discarded, and repeated attention over the same
prompt. We present \method{}, a model-native output interface that stores
the identifier trie as validated GPU tensors. Compared with a Python custom
logit processor (CLP), it reduces CPU--GPU coordination and memory traffic
by avoiding Python constraint callbacks and vocabulary-wide candidate buffers.
It scores compact codebook
candidates and streams the full-vocabulary normalizer, preserving the
original model's valid-token scores in real arithmetic without retraining.
Shared-context attention complements this output path by reusing prompt
key/value blocks across beams while preserving their private decode tails.
We implement both mechanisms in SGLang and evaluate candidate sourcing for
LinkedIn Hiring Assistant and qualification retrieval for the
Qualification Retrieval Platform. The combined system improves
closed-loop throughput by up to 5.57\(\times\) over the CLP baseline and up
to 2.63\(\times\) over BF16 FlashRec at a common beam width of 200 on the
Qualification Retrieval Platform workload.
\end{abstract}
]

\printAffiliationsAndNotice{}

\section{Introduction}

\begin{figure*}[!t]
\centering
\includegraphics[width=\textwidth]{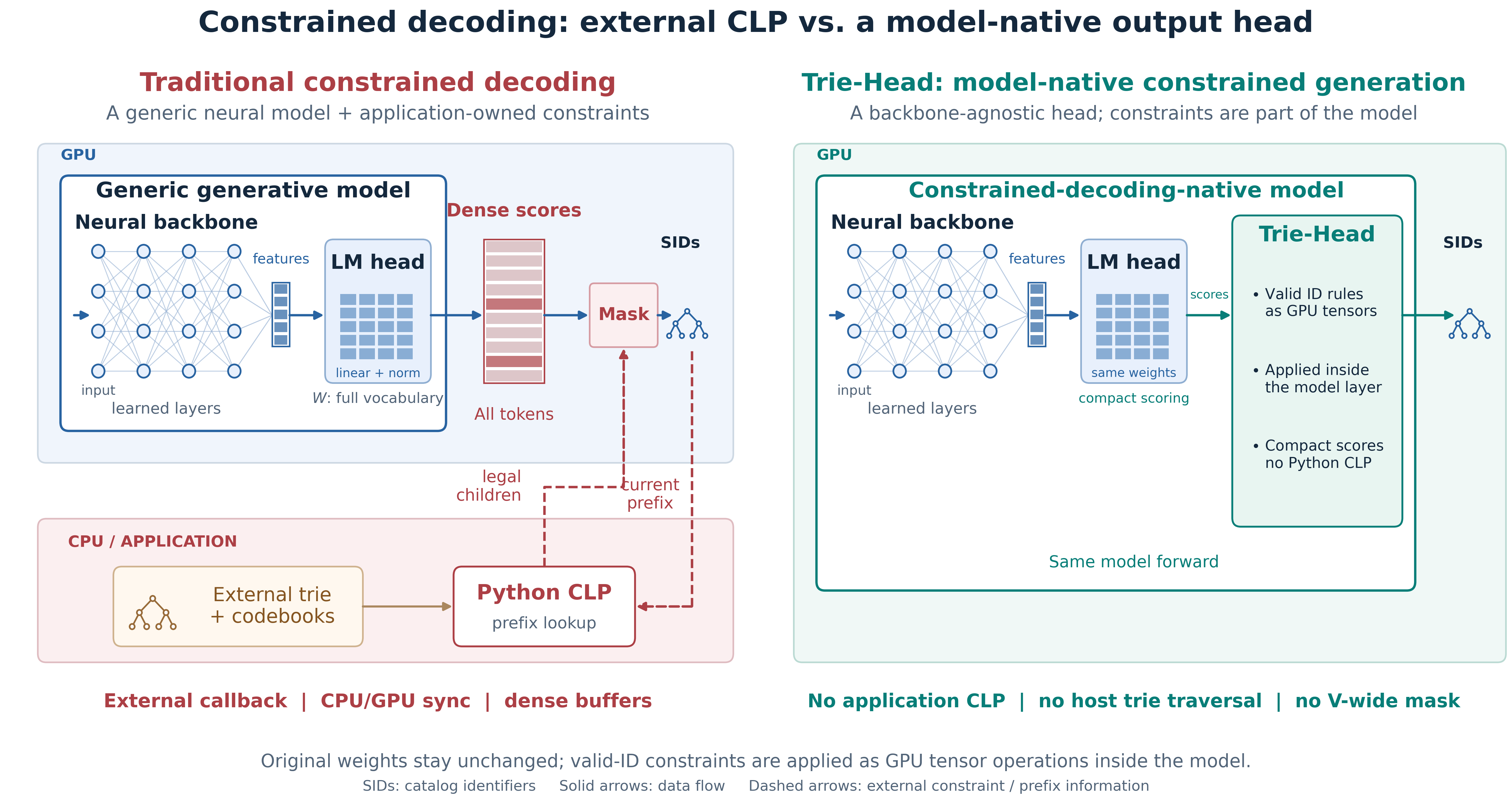}
\caption{External CLP and model-native constrained-decoding architectures.}
\label{fig:model-architecture}
\end{figure*}

A retrieval system must turn a request into a useful candidate pool from a
much larger catalog. Embedding-based retrieval (EBR) encodes the request and
catalog entries as vectors, then searches an index for entries with high
vector similarity. This makes large catalogs searchable efficiently, but
the retrieval score depends on how well that similarity captures the
request's requirements.

\begin{figure*}[!t]
\centering
\begin{minipage}[t]{0.44\textwidth}
\centering
\includegraphics[width=\linewidth]{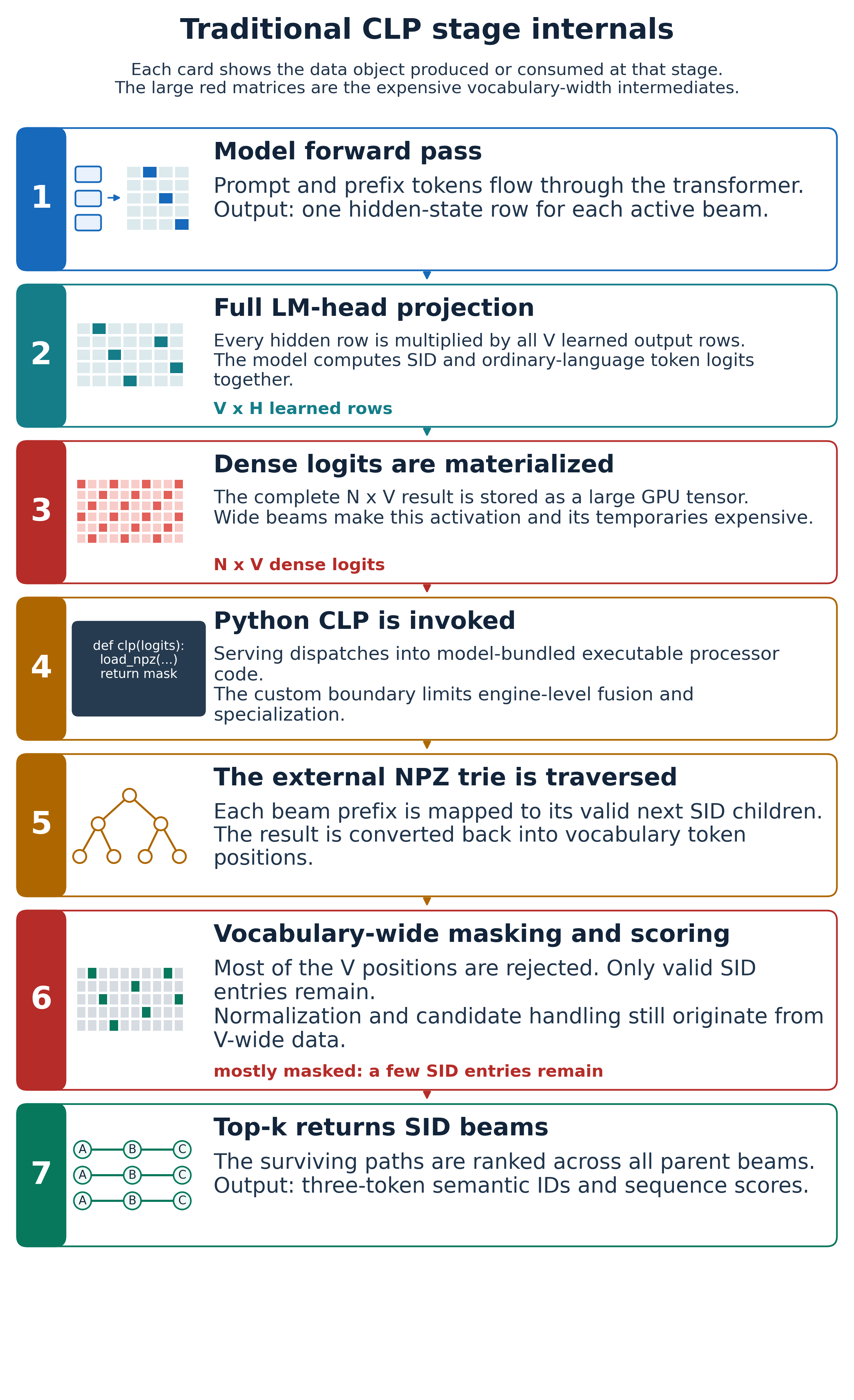}
\end{minipage}
\hfill
\begin{minipage}[t]{0.44\textwidth}
\centering
\includegraphics[width=\linewidth]{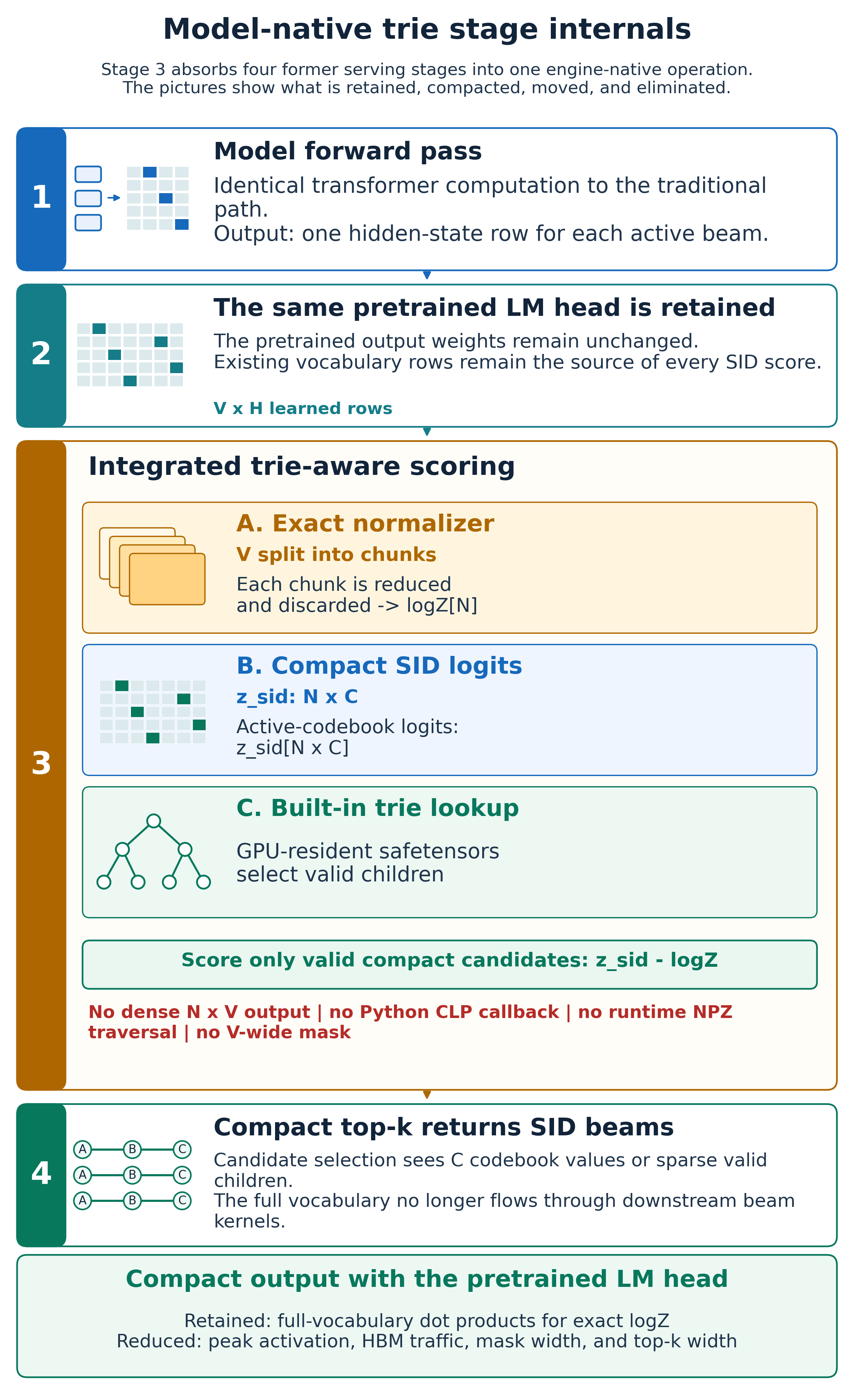}
\end{minipage}
\caption{Baseline and \method{} output-stage dataflows. The baseline
materializes dense logits, crosses a Python CLP boundary, traverses an
externally packaged trie, and applies a vocabulary-wide mask. \method{} keeps
the pretrained LM head and full-vocabulary normalization but packages the trie
with the model and ranks compact valid candidates inside the engine.
\(V\) is vocabulary size and \(H\) is hidden size.}
\label{fig:architecture}
\end{figure*}

Generative retrieval (GR) offers a different way to connect a request to
relevant entries: a model generates their identifiers directly
\cite{tay2022dsi,wang2023nci,decao2021genre}. Rather than comparing separately
encoded vectors, it conditions each identifier token on the request and the
tokens already generated. The model can therefore learn which combinations
of requirements point to which entries, without reducing that decision to
a fixed vector-similarity score. This flexibility makes GR useful as a
complementary candidate source for requests that combine several skills,
preferences, or other contextual requirements.

To make identifier generation practical at catalog scale, semantic
identifiers (SIDs) organize the output space into short, structured
sequences. An SID is a sequence of codes derived from an entity
representation, with each code drawn from a finite token set called a
codebook. Related representations can share code prefixes, allowing the
generator to reuse structure across catalog entries
\cite{sun2023tokenize,rajput2023generative}. A prefix trie organizes the
catalog's identifiers as paths: each edge adds a code, and complete paths
correspond to indexed entries. Generation searches these paths; the returned
SIDs are then resolved to records for downstream ranking.

This structure also makes catalog validity a decoding constraint.
A token can belong to a codebook without extending the current
prefix toward any valid SID. Constrained decoding therefore permits only
the next tokens allowed by the trie. Beam search retains several promising
prefixes at each position to build a candidate pool; its width \(B\)
determines how many paths are kept. Retrieving thousands of candidates means
scoring thousands of partial paths, even when each identifier contains only
three tokens.

The usual implementation places a Python custom logit processor (CLP)
between the model's next-token scores and beam selection. For each partial
identifier, the CLP looks up its position in the trie and asks which next
tokens can still lead to a catalog entry. It leaves those scores unchanged
and sets every other score to negative infinity, preventing beam search
from following an invalid branch.

At wide beams, this simple validity check becomes part of an expensive
inference workload. A standard language-model (LM) head first produces scores
for the full vocabulary, so the CLP uses catalog structure only after
dense scoring has materialized vocabulary-wide tensors. Meanwhile, every
beam attends to the same request context. Wider candidate pools multiply
both costs: output processing that discards most vocabulary entries and
attention that repeatedly reads the shared prompt.

We make the valid output structure part of model execution rather than an
external callback. \method{} implements this
\emph{constrained-decoding-native} output interface by packaging the complete
SID trie as validated GPU tensors alongside the pretrained output weights. Its
interface is backbone-agnostic: it operates on hidden states and output
weights, not the internals of a particular model family. Within the model's
output path, it projects compact codebook candidates, applies trie
membership, and streams the full-vocabulary normalizer. The model therefore
retains its original valid-token scores in real arithmetic without retaining
vocabulary-wide candidate buffers or invoking Python constraint callbacks.
No retraining is required. Shared-context attention complements this output
design by reusing prompt key/value blocks across beams while retaining each
beam's private decode tail.

Two LinkedIn applications make these requirements concrete. LinkedIn
Hiring Assistant helps recruiters source candidates from
natural-language hiring requirements. Its GR component maps combinations of
skills, experience, and preferences to member SIDs, complementing existing
sourcing strategies before ranking and recruiter review. The
Qualification Retrieval Platform starts from member profiles and
retrieves candidate qualifications, such as skills and experience, from a
supplied catalog. GR lets the profile text condition which qualification
identifiers are generated. These applications expose complementary serving
pressures:
LinkedIn Hiring Assistant needs low-latency candidate sourcing, while the
offline Qualification Retrieval Platform emphasizes corpus-scale throughput.

We implement the design in SGLang and compare two serving systems: a
CLP-based baseline and \method{} with shared-context attention. The evaluation
uses both workloads to examine throughput, tail latency, memory, and output
agreement. Closed-loop tests characterize completion capacity; open-loop
replays test whether the system can keep pace with arrivals under a latency
objective. The combined system improves matched closed-loop throughput by
up to 5.57\(\times\), while the external comparison evaluates a common
beam-width configuration with FlashRec \cite{wang2026flashrec}. We also
compare configurations that enable each mechanism separately.

We make three contributions:
\begin{itemize}[leftmargin=*,nosep]
    \item a model-native trie artifact and exact, compact output path.
    \item shared-context attention for wide beams with private decode tails.
    \item an integrated SGLang system evaluated for output validity,
    numerical agreement, throughput, latency, and memory.
\end{itemize}

\section{Workload and Baseline}

\begin{figure*}[!t]
\centering
\includegraphics[width=0.88\textwidth]{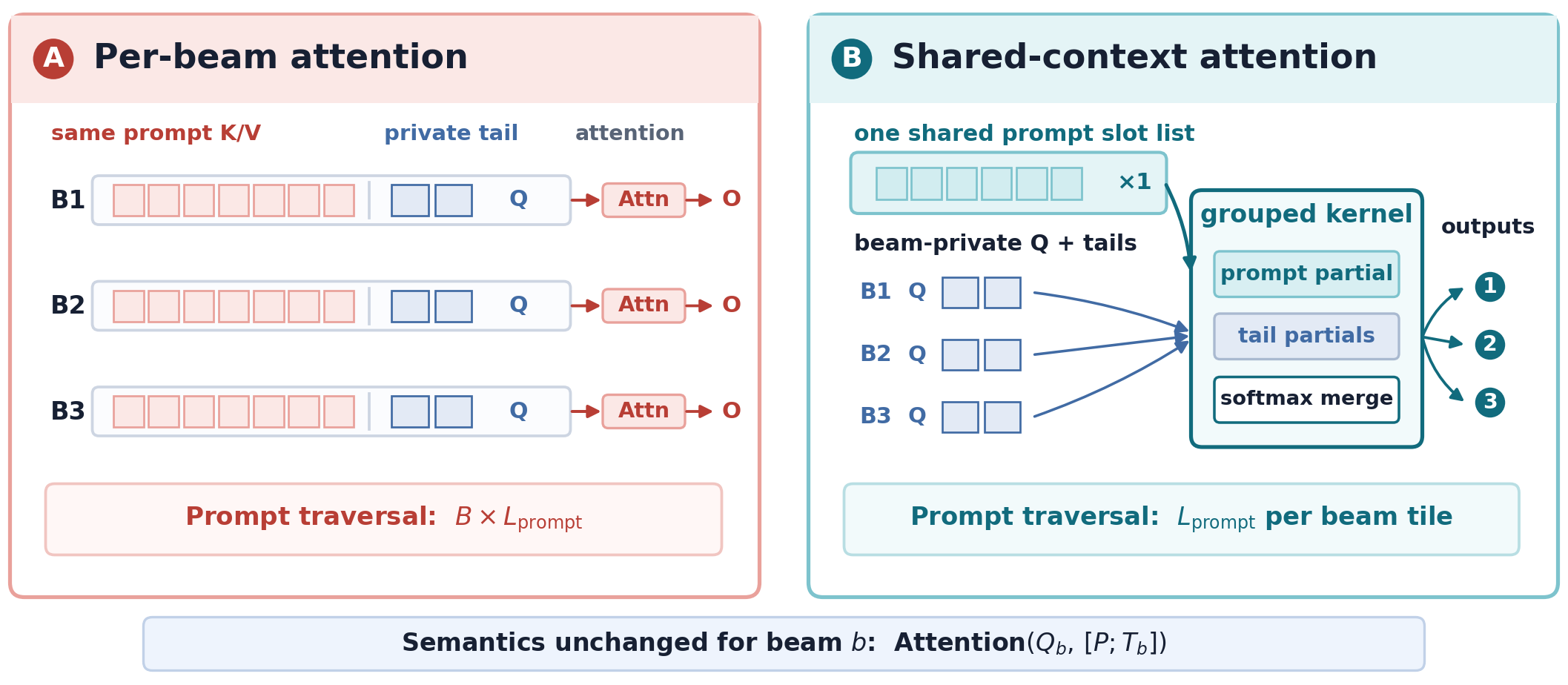}
\caption{Wide retrieval beams share the complete prompt prefix. Standard
row-wise attention traverses the same prompt for every beam. Shared-context
attention reuses prompt K/V blocks across a beam tile and combines the prompt
and each beam's private-tail contributions with an online softmax merge.
\(Q_b\) is beam \(b\)'s query, \(P\) is prompt K/V, and \(T_b\) is its
private-tail K/V. Brackets denote concatenation, and \(O\) is the attention
output. \(L_{\mathrm{prompt}}\), written \(L_p\) in the text, is prompt length.}
\label{fig:shared-attention}
\end{figure*}

\subsection{Candidate sourcing for LinkedIn Hiring Assistant}
\label{sec:hiring-assistant}

LinkedIn Hiring Assistant helps recruiters turn hiring requirements into a
pool of candidates to review. A recruiter describes a role and its required
and preferred qualifications, then refines those requirements during
calibration. The resulting hiring intent is the retrieval model's input:
it may combine role, skills, experience, and location rather than a single
keyword query.

The GR retriever generates member SIDs conditioned on that intent.
SIDs are opaque codes; mapping them to member records occurs in a separate,
access-controlled service.
An application resolver maps the SIDs to member profiles, and the retrieved
candidates join those from existing Boolean and semantic sourcing strategies.
Downstream ranking and recruiter review are separate stages outside the scope
of this work. GR is therefore an additional source
of candidates, not a replacement for ranking or recruiter judgment.
Our workload measures the global-catalog SID-generation stage with a fixed
model checkpoint. Intake, request-specific eligibility constraints,
profile resolution, ranking, and recruiter interactions are outside this
serving replay.

\subsection{Member qualification tagging}
\label{sec:qualification-tagging}

Member qualification tagging identifies which catalog qualifications are
supported by the information a member has provided in their profile.
The labels come from a supplied catalog of
natural-language qualifications spanning skills, experience, education, and
licenses. These labels are model predictions from profile information, not
independently verified qualifications. These predicted labels are not used on
their own to make decisions about members.

The Qualification Retrieval Platform uses GR to generate candidate
qualification SIDs directly from the profile. Its raw SIDs are resolved and deduplicated before request-specific
selection. We take the catalog as given and focus on efficient candidate
generation within member labeling.

\subsection{Shared fixed-depth SID retrieval}

We use three contiguous vocabulary codebooks, each containing \(C\) tokens.
A three-token SID is
\[
    y=(c_0,c_1,c_2), \qquad c_d \in [0,C),
\]
with integer codes \(c_d\). At depth \(d\), code \(c\) maps to
\[
    \tau(d,c) = s + dC + c,
\]
where \(s\) is the first SID token. Only
\(\mathcal{Y}\subseteq[0,C)^3\) names indexed entities. Trie-constrained
decoding retains exactly prefixes that lead to \(\mathcal{Y}\).

\subsection{One request through the baseline}

For an illustrative qualification SID \((17,4,9)\), depth 0 projects the prompt hidden
state against the full LM head, normalizes all \(V\) vocabulary
logits, invokes the Python CLP, queries valid root codes, writes a
vocabulary-width mask, and selects
candidates. The same sequence repeats for children of prefix \((17)\) and then
\((17,4)\). Local candidates from every live row are ranked together to form
the next frontier. A beam-1,800 request repeats this path across up to 1,800
rows. After the first token, those rows also traverse the same profile prompt
while differing only in a one- or two-token tail.
Figures~\ref{fig:model-architecture} and~\ref{fig:architecture} contrast the
architectures and their output-stage dataflows.

\subsection{Two repeated-work bottlenecks}

For \(N\) active rows and vocabulary size \(V\), let \(X\) be the hidden-state
matrix and \(W\) the LM-head weight matrix, with one row per vocabulary token.
The baseline computes and retains
\[
    Z = XW^\top \in \mathbb{R}^{N\times V}
\]
through normalization, CLP dispatch, masking, and top-\(k\). An FP32
intermediate occupies \(4NV\) bytes before masks or selection workspace.
In parallel,
row-wise attention traverses the shared prompt \(N\) times. \method{} removes
the vocabulary-width intermediate. Shared-context attention reuses prompt K/V
blocks while preserving private beam tails.

\section{Trie-Head: Model-Native Trie Decoding}

The design removes repeated work at two points in a decode step. Attention
reuses the prompt across beam queries while keeping their generated tails
separate. The output path then scores the active codebook, computes the
full-vocabulary normalizer, and applies trie membership without retaining a
vocabulary-wide score matrix. Local candidate selection feeds the global
beam frontier. Neither mechanism changes the model weights or catalog.

\subsection{\method{} artifact and validation}

\method{} packages the trie with the pretrained model as a versioned tensor
artifact. Codebook metadata identifies the SID token ranges, and CSR-style
root, prefix, offset, and terminal tensors encode valid paths. Construction
deduplicates and sorts the input identifiers, then reconstructs the stored
paths to check exact set equality. At load time, SGLang checks the schema,
token ranges, sorted groups, and monotonic offsets before moving the tensors
to the GPU. Models that do not opt in retain their ordinary output path.
Opted-in models require a valid artifact; malformed artifacts fail startup
rather than silently changing constraints.

\subsection{\method{} score decomposition}

Let \(\mathcal{V}\) be the vocabulary token set. For hidden state \(h\),
LM-head row \(w_v\), and any model-specific logit transformation \(g\), define
\[
    z_v = g(h^\top w_v), \qquad
    \log p(v\mid h) = z_v - \log\sum_{u\in\mathcal{V}}e^{z_u}.
\]
A dense CLP computes all \(z_v\), normalizes over \(\mathcal{V}\), masks invalid tokens, and selects from the remaining scores.

\method{} projects the active codebook for candidate logits but retains the
full-vocabulary normalizer: \(\log Z\) differs by row and therefore affects
global beam ranking. It scans the LM head in chunks \(\mathcal{V}_j\).
For each chunk, it computes
\[
    m_j = \max_{u\in\mathcal{V}_j} z_u,\qquad
    r_j = \log\sum_{u\in\mathcal{V}_j}e^{z_u-m_j}.
\]
Two partial reductions combine as
\[
\begin{aligned}
    m &= \max(m_a,m_b),\\
    r &= \log\left(e^{r_a+m_a-m}+e^{r_b+m_b-m}\right).
\end{aligned}
\]
After all chunks, \(\log Z=m+r\), and valid token \(t\) receives
\(z_t-\log Z\). Each vocabulary chunk is reduced and discarded immediately.
The scan preserves the model's logit scale and soft cap.

\subsection{\method{} GPU constraint representations}

Depths 0 and 1 use Boolean masks of size \(C\) and \(C^2\). A dense depth-2
mask grows as \(C^3\), so it is appropriate only when that state fits the
memory budget. Otherwise, sorted prefix keys and CSR offsets identify valid
terminal codes without allocating the full codebook product. A GPU kernel
applies the resulting membership test directly to compact candidate scores.
The choice changes storage and lookup cost, not the accepted SID set.

\subsection{Active-codebook grouped projection}
\label{sec:grouped-projection}

The grouped path stably partitions mixed-depth rows, multiplies each group by
its active codebook with an SM90 BF16 grouped GEMM, and restores row order. A
small-batch partition runs within one block; larger batches use count,
prefix-offset, and scatter stages. Unsupported devices or layouts use the
compact projection. Projecting only the active codebook avoids computing
candidates from the other SID depths. Stable partitioning preserves the beam
coordinator's row semantics, and both paths support CUDA graphs.

\subsection{Candidate selection with \texorpdfstring{\(B\)}{B} global candidates}

Generic beam search keeps up to \(2B\) candidates because stop tokens may
consume slots without advancing the frontier. Nonterminal SID depths contain
no stop token, so the global top \(B\) valid extensions are exactly the next
frontier. Retaining \(\min(B,C)\) candidates per row is sufficient: anything
below local rank \(B\) already has \(B\) higher-scoring candidates from its own
row.

\section{Shared-Context Attention}

The attention path stores one prompt-slot list and a beam-major matrix of
private decode slots. Its Triton kernel tiles beam queries, reuses prompt K/V
blocks, and merges the prompt and tail partial softmax states. For prompt
\(c\) and decode tail \(d\), let \((m_c,z_c,o_c)\) and \((m_d,z_d,o_d)\) be
their partial states. Each state contains the maximum attention logit,
the sum of exponentials shifted by that maximum, and the value-weighted sum
using those same unnormalized weights. The merge computes
\[
\begin{aligned}
m &= \max(m_c,m_d),\\
z &= e^{m_c-m}z_c + e^{m_d-m}z_d,\\
o &= e^{m_c-m}o_c + e^{m_d-m}o_d.
\end{aligned}
\]
The final output \(o/z\) is exactly attention over the concatenated prompt and
tail in real arithmetic.

The kernel reads SGLang's native KV slot maps. With the compact output path,
its bounded prompt and tail metadata also support decode CUDA graphs.

\section{SGLang Integration and Correctness}

\subsection{SGLang integration and fallbacks}
\label{sec:serving-integration}

\method{} integrates with model loading, logits processing, beam search,
scheduling, and CUDA graphs. Trie depth and prefix state stay on the GPU.
The scheduler records one trie depth per beam row and gathers prefix state with
the selected parents, avoiding host-side trie reconstruction between steps.
CUDA graphs preallocate bounded prompt, tail, partition, and selection
metadata. Features that alter output semantics---including embedding bias,
unsupported quantization, or LoRA mutation---retain the dense LM head.
Ordinary beam search retains \(2B\) candidates. Unsupported attention layouts
use ordinary attention, while batches outside the captured graph shapes
execute eagerly.

Beam width is a per-request parameter: the same server accepts small and large
beam requests without a restart. Mixed widths share the native scheduler and
memory pools, while each request keeps its own frontier, result count, and
trie prefix state. This separates deployment from the choice of retrieval-pool
size. Available sequence rows and KV capacity still bound admission.

\subsection{Complexity and memory effects}

\method{} preserves full-vocabulary projection arithmetic for exact
normalization, but changes what survives that computation. The dense path
materializes and retains \(N\times V\) scores through normalization, masking,
and top-\(k\). The model-native path reduces each vocabulary chunk immediately
and retains \(O(N)\) normalizer state plus \(N\times C\) compact candidate
scores. Avoiding the retained dense intermediate removes \(4NV\) bytes for
FP32 scores before accounting for masks and selection workspace.
Candidate storage is bounded by the active codebook, and sparse trie membership
scales with realized prefixes and paths rather than \(C^3\).

For attention, a prompt of length \(L_p\) is logically shared by all \(B\)
beams. Row-wise decoding loads prompt K/V blocks for each beam, whereas the grouped
kernel loads them once per query tile and retains per-beam work only for the
short private tail. The optimization therefore grows with prompt length and
beam width without changing the attention result. Finally, fixed-length
selection reduces the global candidate pool from \(2B\) to \(B\) while local
retention remains bounded by \(\min(B,C)\). The three reductions target
different resources: output-intermediate memory and masking, prompt K/V
traffic, and global selection width.

\subsection{Correctness guarantees}
\label{sec:design-guarantees}

\paragraph{Lossless representation.}
The artifact reconstructs exactly the deduplicated canonical SID set
\(\mathcal{Y}\). Sorted roots own contiguous second-level slices, each prefix
owns a contiguous terminal slice, and construction emits one entry per unique
triple and none outside \(\mathcal{Y}\). Range checks precede integer
narrowing, preserving values across storage widths. Following the two offset
arrays therefore enumerates the same triples and cardinality as the source.
For a fixed prefix, trie membership is deterministic: its valid child set is
fixed by the artifact, independent of batch scheduling and floating-point
reductions.

\paragraph{Lossless scoring in real arithmetic.}
For every valid child, \method{} and dense full-vocabulary scoring followed by
trie masking assign the same log-probability in real arithmetic. They use the
same hidden state, LM-head row, scaling, and soft cap. The streamed recurrence
computes the same full-vocabulary log-sum-exp, and masking removes only invalid
entries. Subtracting the same \(\log Z\) from the same candidate logit yields
the same score, including comparisons between candidates from different parent
rows.

\paragraph{Attention.}
Shared-context attention equals attention over the concatenated prompt and
private tail in real arithmetic. Each pass computes a partial maximum,
exponential sum, and value-weighted sum. Rescaling to the joint maximum and
adding reconstructs the exact softmax numerator and denominator. Their quotient
is therefore unchanged for every query row.

\paragraph{Search.}
At a nonterminal SID step, local top-\(\min(B,C)\) followed by global top-\(B\)
returns the dense masked-search frontier. No candidate terminates, and anything
below local rank \(B\) already has \(B\) higher-scoring candidates from its own
row, so it cannot enter the global top \(B\). When \(B>C\), retaining all
\(C\) children remains sufficient. Tied boundary candidates require the same
deterministic total order for identity. Finite-precision kernels can reorder
close scores: deterministic membership does not imply bitwise-identical beam
outputs. \supplementref{app:agreement} reports the resulting set agreement.

\section{Evaluation}
\label{sec:evaluation}

\subsection{Experimental methodology}

We compare two serving systems. The \emph{CLP baseline} uses SGLang 0.5.19's
native beam search with dense vocabulary scoring and a Python trie constraint.
The CLP uses a preloaded prefix index and batched masking.
The \emph{combined system} uses \method{} and shared-context attention.
The comparison holds model weights, tokenizer configuration, prompt replay,
beam width, and model-specific resource budgets fixed.
Both systems use SGLang's native beam coordinator; compatible attention
paths use FlashInfer \cite{ye2025flashinfer}.

\begin{table}[t]
\centering
\small
\caption{Production-derived workloads. \(B\) is both beam width and raw SID
result count. Prompt lengths are mean tokens per measured prompt, rounded
to the nearest token.}
\label{tab:workloads}
\resizebox{\columnwidth}{!}{%
\begin{tabular}{lll}
\toprule
Workload & \shortstack[l]{Qualification\\Retrieval Platform} &
\shortstack[l]{LinkedIn Hiring\\Assistant} \\
\midrule
Nominal model scale & 0.6B & 1.7B \\
Beam width \(B\) & 1,800 & 2,000 \\
Codebook width \(C\) & 256 & 8,192 \\
Trie representation & Dense & Sparse \\
Avg. tokens per prompt & 1,653 & 252 \\
\bottomrule
\end{tabular}
}
\end{table}

\paragraph{Workloads and platform.}
Table~\ref{tab:workloads} summarizes the two production-derived replays.
The Qualification Retrieval Platform exercises dense trie membership and
longer contexts; LinkedIn Hiring Assistant exercises sparse membership and
wider codebooks. Every run uses one NVIDIA H200 with BF16 weights and KV cache,
CUDA graphs, and a fixed sequence-row budget. Prefix caching is enabled for
the Qualification Retrieval Platform and disabled for LinkedIn Hiring Assistant.
Requests queue when the resident beam groups
exhaust the row budget, so client concurrency can exceed GPU-resident
concurrency. Replays use de-identified request data and ran only on
LinkedIn-controlled infrastructure, including the external-engine comparisons
in Sections~\ref{sec:flashrec} and~\ref{sec:external-gr}.

\paragraph{Measurement.}
We time asynchronous engine submission through the complete raw beam response.
This includes prefill, decoding, constraints, attention, selection, and internal
response assembly, but excludes input tokenization, external HTTP transport,
SID resolution, downstream ranking, and file writing. Memory is the maximum
sampled device usage over the benchmark process, not an isolated operator
allocation.

Each point uses five independent processes, each replaying 1,000 measured
requests after workload-specific warmup. Closed-loop tests hold client
concurrency fixed. Open-loop tests submit uncapped Poisson arrivals with
fixed per-run seeds. We report means with Student's \(t\) confidence intervals
and use independent run-level bootstrap resampling for system ratios.
\supplementref{app:reproducibility} gives the resource and warmup settings.

\paragraph{Validation.}
\label{sec:output-validation}
We check result count, uniqueness, and trie membership for every response
outside the inference timer. An independent dense-scoring oracle tests the
compact output path on synthetic tries, including exact ties and beam widths
larger than a codebook. We separately measure finite-precision SID agreement:
accumulation order can change near-tied scores even when the real-arithmetic
formulas agree \cite{he2025nondeterminism}.
Table~\ref{tab:agreement} reports the measured agreement.
Full-beam and top-100 SID overlap are averaged over 1,000 prompts per
workload; resolved-set overlap uses the same application resolver on 50
prompts. The results support a distinction between exact constraints and
finite-precision execution: every output is valid, but the ranked SID sets
are not identical. They do not establish unchanged retrieval relevance.
Any production change to the serving path is subject to LinkedIn's standard
quality and responsible-AI evaluation before launch.
\supplementref{app:agreement} gives the complete validation protocol.

\begin{table*}[t]
\centering
\small
\caption{CLP versus combined set agreement (\%). Full-beam and top-100
results use 1,000 prompts; resolved-set results use 50 prompts, with cutoffs
100 for the Qualification Retrieval Platform and 200 for LinkedIn Hiring
Assistant. All measured responses
have valid, unique SIDs and the requested result count. Set agreement is
not a relevance or bitwise-equivalence guarantee.}
\label{tab:agreement}
\begin{tabular}{lrrr}
\toprule
Workload & Full SID beam & Top-100 SIDs & Resolved set \\
\midrule
Qualification Retrieval Platform & 97.50 & 96.12 & 96.10 \\
LinkedIn Hiring Assistant & 93.58 & 91.47 & 91.66 \\
\bottomrule
\end{tabular}
\end{table*}

\subsection{Performance}

\begin{figure*}[t]
\centering
\includegraphics[width=0.86\textwidth]{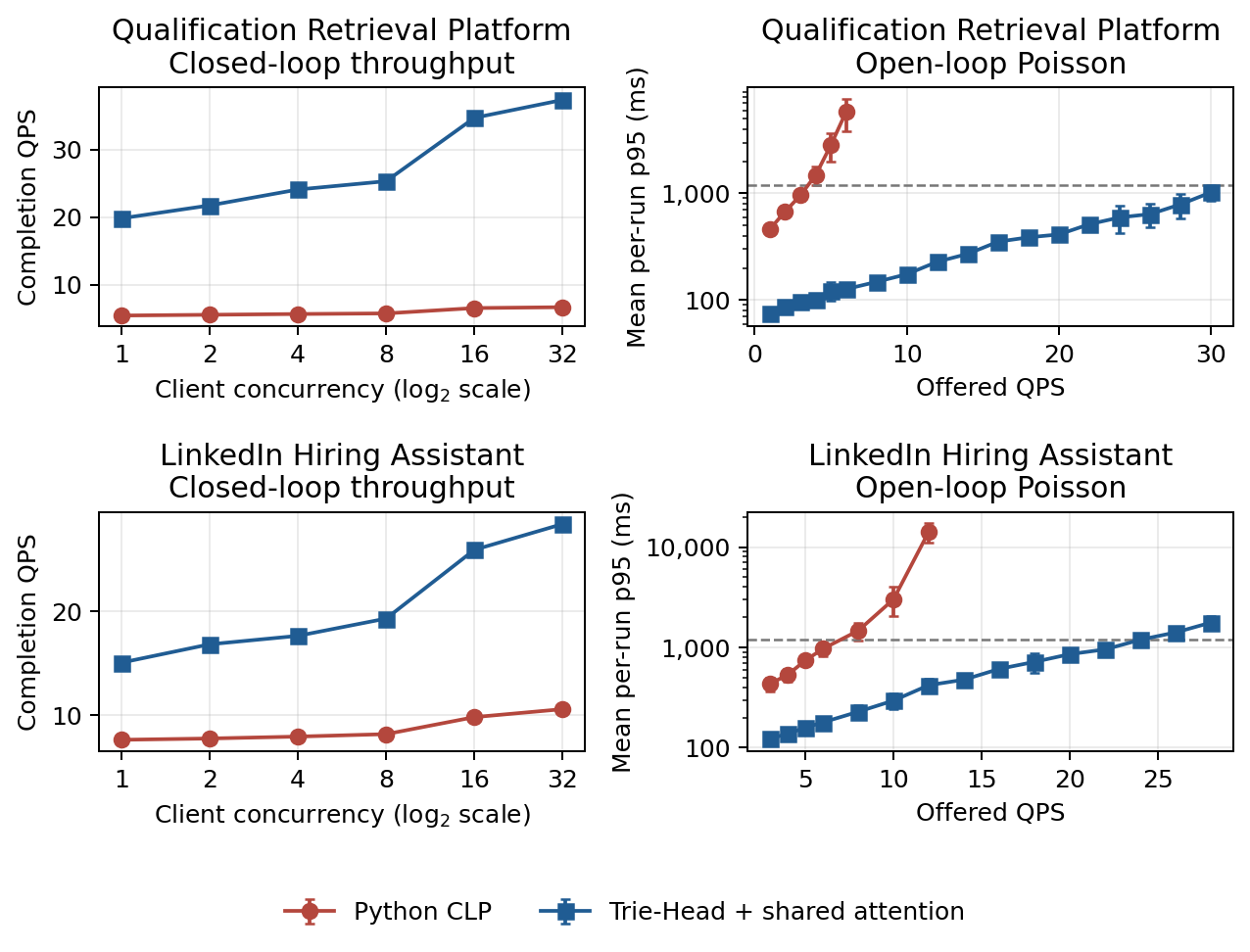}
\caption{CLP baseline versus \method{} with shared-context attention, five repetitions per
point. Latency is mean per-run p95 on a logarithmic axis. Error bars are 95\%
mean confidence intervals. Dashed lines mark the 1.2-second p95 objective.}
\label{fig:end-to-end}
\end{figure*}

The combined system improves throughput at every matched closed-loop point
in Figure~\ref{fig:end-to-end}: 3.63--5.57\(\times\) for the Qualification
Retrieval Platform and
1.98--2.69\(\times\) for LinkedIn Hiring Assistant. The open-loop panels show a corresponding
increase in the arrival rate that can be served before queueing raises tail
latency. Device memory does not decrease uniformly: LinkedIn Hiring Assistant uses more memory
at high concurrency while completing more requests.
\supplementref{app:sweeps} gives the complete throughput, latency, and memory
grids.

\begin{table*}[t]
\centering
\small
\caption{Highest measured offered load (QPS) with mean per-run p95 at most 1.2~s
and valid completion QPS at least 98\% of offered QPS.
Latency pairs are CLP / combined.}
\label{tab:sustainable}
\begin{tabular}{lrrrr}
\toprule
Model & CLP & Combined & Increase & p95 (ms) \\
\midrule
Qualification Retrieval Platform & 3 & 28 & 9.33\(\times\) & 959 / 791 \\
LinkedIn Hiring Assistant & 6 & 24 & 4.00\(\times\) & 968 / 1,194 \\
\bottomrule
\end{tabular}
\end{table*}

Table~\ref{tab:sustainable} reports the highest measured offered rate meeting
both the latency and completion criteria. The Qualification Retrieval
Platform supports 28 offered QPS
instead of 3, and LinkedIn Hiring Assistant supports 24 instead of 6. These are offered loads,
not exact completion rates: the combined system completes 27.44 and
23.54 valid requests per second, respectively. The Qualification Retrieval
Platform also meets the latency
threshold at 30 offered QPS, but its completion rate falls below the 98\%
requirement. All measured responses contain the requested number of unique,
trie-valid SIDs.

\subsection{Component analysis}
\label{sec:components}

Table~\ref{tab:components} compares \method{} alone, shared-context attention alone, and the combined
system against the same SGLang 0.5.19 CLP baseline. At client concurrency 4
for the Qualification Retrieval Platform, the standalone configurations reach 1.21\(\times\) and
2.52\(\times\) baseline throughput; at concurrency 8 for LinkedIn Hiring Assistant, both reach
1.50\(\times\). The combined system reaches 4.23\(\times\) and
2.37\(\times\), respectively, at these operating points.

\begin{table*}[t]
\centering
\small
\caption{Component serving comparison at client concurrency 4 for the
Qualification Retrieval Platform and 8 for LinkedIn Hiring Assistant.
QPS is the five-run mean with a 95\% mean confidence interval.
Every gain is relative to SGLang 0.5.19 CLP. Source and execution settings
differ across rows; the ratios are not a same-build causal ablation.}
\label{tab:components}
\begin{tabular}{lrrrr}
\toprule
Configuration & \multicolumn{2}{c}{\shortstack{Qualification Retrieval\\Platform}} &
\multicolumn{2}{c}{\shortstack{LinkedIn Hiring\\Assistant}} \\
\cmidrule(lr){2-3}\cmidrule(lr){4-5}
 & QPS & Gain & QPS & Gain \\
\midrule
CLP baseline & \(5.691 \pm 0.015\) & 1.00\(\times\) &
\(8.142 \pm 0.050\) & 1.00\(\times\) \\
\method{} only & \(6.873 \pm 0.011\) & 1.21\(\times\) &
\(12.200 \pm 0.062\) & 1.50\(\times\) \\
Shared attention only & \(14.355 \pm 0.220\) & 2.52\(\times\) &
\(12.251 \pm 0.060\) & 1.50\(\times\) \\
\method{} + shared attention & \(24.073 \pm 0.305\) & 4.23\(\times\) &
\(19.286 \pm 0.186\) & 2.37\(\times\) \\
\bottomrule
\end{tabular}
\end{table*}

The baseline and combined measurements are the corresponding points from
the primary serving comparison. The standalone measurements use separate
source builds, and shared-attention-only decode executes eagerly.
These results compare serving configurations, not isolated causal
contributions. Fixed-concurrency throughput also differs from the
open-loop offered-load increases in Table~\ref{tab:sustainable}.
\supplementref{app:components} gives the measurements and execution settings.

\subsection{External comparison with FlashRec}
\label{sec:flashrec}

We evaluate FlashRec \cite{wang2026flashrec} on the Qualification Retrieval
Platform replay at a common beam width supported by both systems. Each run uses one H200, BF16 weights
and KV cache, prefix caching, and CUDA graphs. We hold the model, input tokens,
and complete catalog fixed and vary client concurrency. The combined system
has five repetitions per point and FlashRec has two; the figure therefore
shows confidence intervals only for the combined system.

\begin{figure*}[t]
\centering
\includegraphics[width=0.96\textwidth]{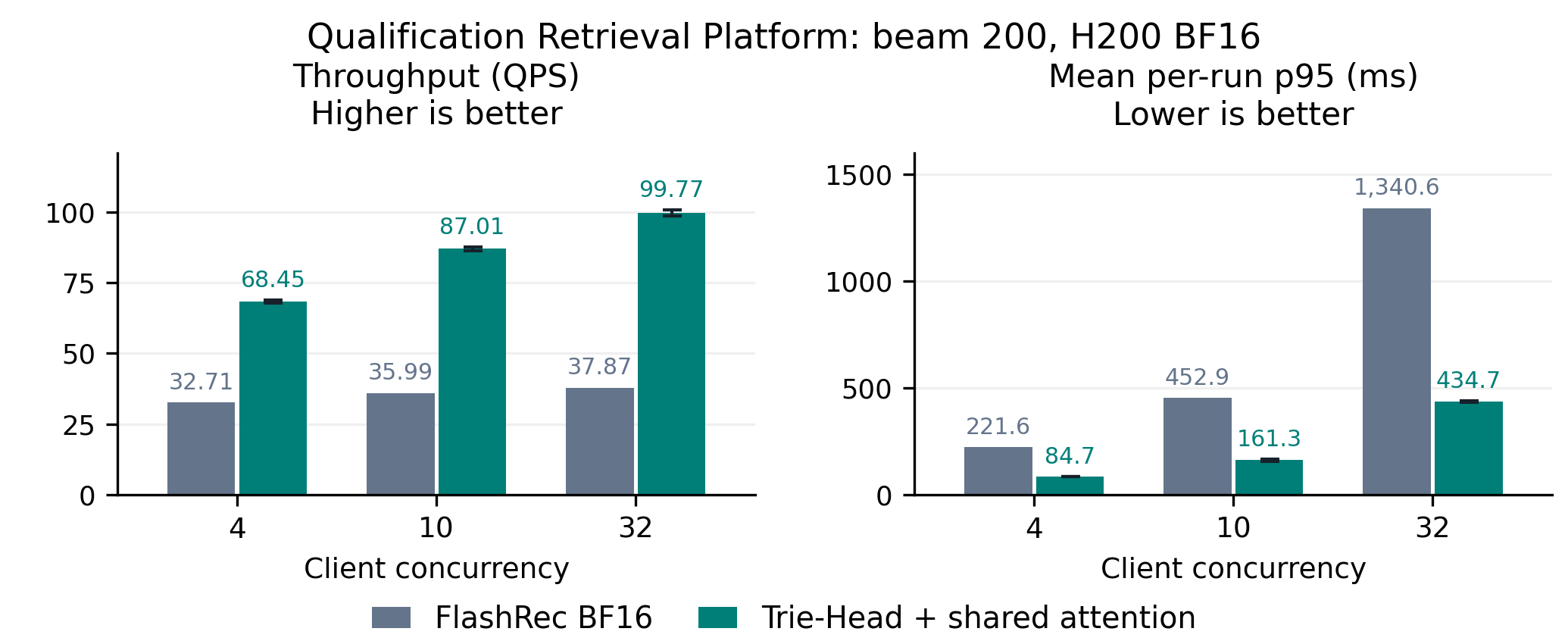}
\caption{External Qualification Retrieval Platform comparison at beam 200. Bars show
mean QPS (left) and mean per-run p95 latency (right): two FlashRec
runs and five combined runs at each concurrency. Error bars are 95\%
mean intervals for the combined system only. Each run uses one H200 with BF16.}
\label{fig:flashrec}
\end{figure*}

The combined system improves both completion throughput and tail latency
across the tested concurrencies (Figure~\ref{fig:flashrec}). This is a serving
comparison, not a score-equivalence claim: our path retains the full-vocabulary
normalizer, whereas FlashRec uses its active-codebook normalizer.

The evaluated FlashRec native path also imposes compatibility bounds. Its
fused selector supports beam widths up to 512, below the production beam
requirements of both workloads. The Qualification Retrieval Platform also
exceeds the root initializer's
supported beam/codebook relation. These bounds concern the native fast path,
not every possible engine fallback.

\subsection{Compatible-catalog comparison with GR-Inference}
\label{sec:external-gr}

We also evaluate an external NVIDIA-supplied GR-Inference prototype.
Its native REC path supports beam widths up to 1,024 and fan-out up to
4,096; LinkedIn Hiring Assistant exceeds both limits. The path requires candidate-normalized
scores and therefore cannot execute the full-vocabulary primary protocol.
We construct a compatible comparison by deterministically limiting fan-out
to the supported bound and using beam 1,000. The restriction is based on trie
ordering, not model scores or observed performance. Both systems use the same
model, inputs, restricted catalog, and candidate-normalized scoring on one
H200 with BF16 weights and KV cache. The primary CLP comparison continues to
use the complete catalog and full-vocabulary normalization.

\begin{table}[t]
\centering
\small
\caption{Compatibility-limited LinkedIn Hiring Assistant comparison. QPS is the mean completed
request rate; p95 values are the two per-run measurements.}
\label{tab:external-gr}
\begin{tabular}{lrr}
\toprule
System & QPS & p95 latency (ms) \\
\midrule
GR-Inference & 58.42 & 1,312 / 1,630 \\
\method{} + shared attention & \textbf{61.07} & \textbf{794 / 1,130} \\
\bottomrule
\end{tabular}
\end{table}

Table~\ref{tab:external-gr} shows an advantage at the tested compatible
operating point. It does not establish production-catalog throughput or
ranked-output equivalence between engines.

\section{Related Work}

Generative retrieval learns document or entity identifiers
\cite{tay2022dsi,wang2023nci,decao2021genre,sun2023tokenize,rajput2023generative}.
Constrained beam search, NeuroLogic, grammar decoders, and XGrammar support
lexical, logical, and structured constraints
\cite{anderson2017constrained,lu2021neurologic,geng2023grammar,dong2025xgrammar}.
STATIC vectorizes SID-trie traversal and masking \cite{su2026static}.
\method{} combines compact projection, exact full-vocabulary normalization,
GPU membership, and candidate retention for fixed-depth SID tries.
Table~\ref{tab:mechanisms} summarizes the complementary execution mechanisms.

\begin{table}[t]
\centering
\small
\caption{Qualitative mechanism comparison.}
\label{tab:mechanisms}
\begin{tabular}{lp{0.61\columnwidth}}
\toprule
System & Primary mechanism \\
\midrule
STATIC & Vectorized trie transitions and masking \\
XGrammar & Grammar-aware token checks and masks \\
Hydragen & Exact shared-prefix attention \\
xGR & Staged attention, sorting, and scheduling \\
This work & Model-native compact trie scoring and shared-context beam attention \\
\bottomrule
\end{tabular}
\end{table}

PagedAttention and SGLang improve KV-cache management, batching, scheduling,
and cache reuse \cite{kwon2023pagedattention,zheng2024sglang}.
Hydragen separates shared-prefix and unique-suffix
attention and merges their softmax states exactly
\cite{juravsky2024hydragen}. Our kernel applies that decomposition during
autoregressive SID decoding across thousands of beams using SGLang's native
slot maps and CUDA-graph shapes. xGR stages shared-prefix and beam-tail
attention, accelerates sorting, and overlaps execution \cite{sun2025xgr}.
Our attention integration complements \method{}'s model-native output
contract by reusing prompt K/V blocks across beam queries within each tile.

Multiple-Item Scoring (MIS), introduced for online ranking in MixLM
\cite{li2026mixlm} and used in Semantic Search at LinkedIn
\cite{borisyuk2026semanticsearch}, shares query-prefix computation across
candidate items while masking inter-item attention. It scores a supplied
candidate set in a prefill pass, whereas our shared-context attention reuses
prompt K/V during autoregressive SID beam expansion.

Adaptive softmax changes the output parameterization
\cite{grave2017adaptive}, while vocabulary selection approximates
normalization over a subset \cite{jean2015vocabulary}. \method{} retains the
pretrained LM head and full-vocabulary normalizer, preserving valid-token
scores and cross-row ranking in real arithmetic. This distinction is essential
for beam search because row-specific normalization constants affect which
parent row contributes the global winners.

Job-understanding systems generate attribute candidates from job text and use
learned selection to identify relevant values
\cite{li2020deepjob,xu2026unifiedsemantic}. Our member qualification-tagging
workloads have a related candidate-generation stage, but take member profiles
as input. We optimize GR serving for that candidate-generation stage.

\section{Discussion and Limitations}

\method{} separates a model-independent logits interface from its serving
integration. The implementation still requires a standard vocabulary
projection, compatible logit transformations, and an artifact matching the
model, tokenizer, and static trie. It supports three-token SIDs in equal-sized,
contiguous codebooks rather than arbitrary grammars or request-dependent constraints.
Shared-context attention additionally requires compatible causal attention
and KV layouts; ordinary attention remains available when it cannot be used.
The measured performance covers these two GR workloads on H200, not all
model families or hardware.

Exact membership and real-arithmetic equivalence do not guarantee identical
finite-precision rankings. SID and resolved-set agreement therefore
complement the structural checks, but cannot establish unchanged relevance
without labeled evaluation. Exact normalization also retains full-vocabulary
projection arithmetic, so the benefit depends on how much candidate
processing and shared-prompt traversal the workload would otherwise repeat.
For offline deployment, tokenization, identifier resolution, and CPU-side
request handling remain part of the workload. Their costs do not fall in
proportion to GPU inference time. A production comparison must therefore
measure complete job wall time and resource use with the same request set,
beam policy, and host allocation; single-GPU QPS alone cannot establish a
pipeline-level reduction.

\section{Conclusion}

Catalog constraints need not remain an external callback.
\method{} makes the valid identifier structure part of model execution,
combining compact candidates with full-model scoring in real arithmetic.
Shared-context attention addresses the complementary cost of traversing the
same prompt across wide beams. Together, the mechanisms improve serving
efficiency without retraining or changing the identifier catalog.
The evaluation measures completion capacity and tail latency separately,
and distinguishes exact catalog validity from finite-precision agreement.

\paragraph{Implementation availability and support.}
The implementation is available as a proposed SGLang contribution in
\href{https://github.com/sgl-project/sglang/pull/44349}
{SGLang pull request \#44349}.
Production data, model checkpoints, and SID catalogs remain proprietary and
are not released. LinkedIn supported the
work.

\bibliographystyle{mlsys2025}
\bibliography{references}

\section*{AI Usage Disclosure}
GitHub Copilot, including Copilot CLI, assisted with manuscript drafting,
organization, and language editing; literature and citation searches;
benchmark and validation tooling; statistical-analysis and figure-generation
code; experiment configuration, execution, and inspection; and submission
policy and formatting checks. Assistance included iterative text and code
suggestions, edits, and command execution, rather than proofreading alone.
Reported performance values are derived from recorded benchmark measurements,
not AI-generated estimates. The authors retain responsibility for the
correctness, originality, attribution, and integrity of the code, experiments,
citations, claims, and final text.

\ifdefined\arxivpreprint
\clearpage
\appendix
\section{Complete Performance Results}
\label{app:sweeps}

This appendix gives the complete serving grids and the component comparison.
The serving grids compare the CLP baseline and \method{} with shared-context
attention. Each cell reports five processes replaying the same measured prompt set.

\subsection{Closed-loop capacity}

Table~\ref{tab:appendix-closed} reports completion rate, mean per-run p95,
and maximum observed device memory. Resource budgets remain fixed as client
concurrency increases. The highest measured rate is a maximum over this grid,
not an estimate of the hardware's absolute capacity.

\begin{table*}[t]
\centering
\small
\caption{Closed-loop comparison. Latency is mean per-run p95 in milliseconds.
Memory is maximum observed device usage across five repetitions, in MiB.}
\label{tab:appendix-closed}
\begin{tabular}{p{2.1cm}rrrrrrr}
\toprule
Model & Concurrency & CLP QPS & Combined QPS &
CLP p95 & Combined p95 & CLP MiB & Combined MiB \\
\midrule
\shortstack[l]{Qualification\\Retrieval\\Platform} & 1 & 5.466 & 19.818 & 307.7 & 60.3 & 116,008 & 111,296 \\
 & 2 & 5.579 & 21.727 & 575.5 & 112.7 & 118,026 & 113,112 \\
 & 4 & 5.691 & 24.073 & 1,093.6 & 201.2 & 122,216 & 117,694 \\
 & 8 & 5.781 & 25.300 & 1,943.5 & 443.8 & 132,506 & 126,914 \\
 & 10 & 6.385 & 28.437 & 2,470.4 & 558.8 & 134,572 & 129,208 \\
 & 16 & 6.571 & 34.660 & 3,720.5 & 855.3 & 132,520 & 129,208 \\
 & 32 & 6.699 & 37.305 & 6,770.3 & 1,452.4 & 134,646 & 129,208 \\
\midrule
\shortstack[l]{LinkedIn\\Hiring\\Assistant} & 1 & 7.596 & 15.035 & 143.2 & 71.6 & 89,718 & 84,402 \\
 & 2 & 7.723 & 16.805 & 289.4 & 131.6 & 92,264 & 86,704 \\
 & 4 & 7.907 & 17.620 & 701.1 & 228.9 & 97,574 & 92,042 \\
 & 8 & 8.142 & 19.286 & 1,370.7 & 648.7 & 108,108 & 108,130 \\
 & 10 & 8.383 & 19.841 & 1,706.8 & 870.2 & 113,528 & 115,530 \\
 & 16 & 9.783 & 25.915 & 2,143.0 & 1,088.8 & 113,528 & 115,530 \\
 & 32 & 10.551 & 28.402 & 3,883.1 & 1,722.6 & 113,496 & 115,530 \\
\bottomrule
\end{tabular}
\end{table*}

\subsection{Component comparison}
\label{app:components}

Table~\ref{tab:appendix-components} compares the standalone and combined
configurations against the SGLang 0.5.19 CLP baseline. CLP and combined rows
are the corresponding measurements in Table~\ref{tab:appendix-closed}.
Head-only includes the complete native trie output path, including membership
and candidate selection, with ordinary attention. Attention-only retains
dense scoring and the model-bundled Python CLP. All configurations use
full-vocabulary normalization, the same model weights and beam widths,
and the workload-specific memory and sequence-row budgets.

\begin{table*}[t]
\centering
\small
\caption{Component serving comparison at client concurrency 4 for the
Qualification Retrieval Platform and 8 for LinkedIn Hiring Assistant.
QPS is the five-run mean with a 95\% mean confidence interval.
Every gain is relative to SGLang 0.5.19 CLP. Source and execution settings
differ across rows; the ratios are not a same-build causal ablation.}
\label{tab:appendix-components}
\begin{tabular}{lrrrr}
\toprule
Configuration & \multicolumn{2}{c}{\shortstack{Qualification Retrieval\\Platform}} &
\multicolumn{2}{c}{\shortstack{LinkedIn Hiring\\Assistant}} \\
\cmidrule(lr){2-3}\cmidrule(lr){4-5}
 & QPS & Gain & QPS & Gain \\
\midrule
CLP baseline & \(5.691 \pm 0.015\) & 1.00\(\times\) &
\(8.142 \pm 0.050\) & 1.00\(\times\) \\
\method{} only & \(6.873 \pm 0.011\) & 1.21\(\times\) &
\(12.200 \pm 0.062\) & 1.50\(\times\) \\
Shared attention only & \(14.355 \pm 0.220\) & 2.52\(\times\) &
\(12.251 \pm 0.060\) & 1.50\(\times\) \\
\method{} + shared attention & \(24.073 \pm 0.305\) & 4.23\(\times\) &
\(19.286 \pm 0.186\) & 2.37\(\times\) \\
\bottomrule
\end{tabular}
\end{table*}

Each row uses five independent processes with 1,000 measured requests after
warmup. Standalone measurements use a source-built SGLang with the 0.5.21
runtime snapshot. Their execution order rotates across repetitions.
Prefill graphs are enabled; shared-attention-only decode executes eagerly
because its graph integration requires compact-head metadata.
Head-only uses supported decode graph paths. For the Qualification Retrieval
Platform it retains ordinary
FlashInfer tensor-core attention, as in the CLP baseline, with a graph-group
limit of 4. An explicit 4~GiB attention workspace supports that capture.
The other standalone configurations use graph-group limits of 9 for the
Qualification Retrieval Platform and 10 for LinkedIn Hiring Assistant.
The primary points capture up to their
respective client concurrencies. Source revisions, input
fingerprints, and backend settings accompany the supporting results.
Diagnostic route checks verify the mechanisms outside performance timing.

These ratios include differences in serving implementation and graph
execution, so they cannot partition the primary system-level gain.
The standalone factors need not multiply to the combined factor.
The fixed-concurrency results also do not decompose LinkedIn Hiring Assistant's
4\(\times\) increase in passing offered load: that result uses the joint
latency and completion criterion, not throughput at a fixed concurrency.

\subsection{Open-loop latency and completion}

Table~\ref{tab:appendix-open} reports the complete offered-load grid.
The joint serving criterion requires mean per-run p95 at most 1,200~ms,
valid completion QPS at least 98\% of offered QPS, and the requested number
of unique, valid SIDs in every response. A point can meet the latency
threshold without meeting the rate threshold; both conditions are required.
The criterion applies to run means, not to a guarantee for every individual
request or repetition.

\begin{table*}[t]
\centering
\small
\caption{Open-loop comparison with Poisson arrivals. Latency is mean per-run
p95 in milliseconds. Dashes indicate loads without a CLP measurement.}
\label{tab:appendix-open}
\begin{tabular}{p{2.1cm}rrrrr}
\toprule
Model & Offered QPS & CLP completion QPS & Combined completion QPS &
CLP p95 & Combined p95 \\
\midrule
\shortstack[l]{Qualification\\Retrieval\\Platform} & 1 & 1.002 & 1.002 & 463.6 & 74.2 \\
 & 2 & 2.002 & 2.003 & 676.5 & 87.4 \\
 & 3 & 3.002 & 3.005 & 959.2 & 96.6 \\
 & 4 & 3.995 & 4.006 & 1,481.5 & 101.9 \\
 & 5 & 4.990 & 5.007 & 2,831.2 & 122.4 \\
 & 6 & 5.856 & 6.008 & 5,806.5 & 126.7 \\
 & 8 & -- & 8.010 & -- & 148.1 \\
 & 10 & -- & 10.010 & -- & 174.8 \\
 & 12 & -- & 12.010 & -- & 228.7 \\
 & 14 & -- & 14.009 & -- & 271.2 \\
 & 16 & -- & 16.008 & -- & 353.0 \\
 & 18 & -- & 18.006 & -- & 387.7 \\
 & 20 & -- & 20.001 & -- & 412.6 \\
 & 22 & -- & 21.998 & -- & 514.7 \\
 & 24 & -- & 23.992 & -- & 596.1 \\
 & 26 & -- & 25.986 & -- & 640.2 \\
 & 28 & -- & 27.444 & -- & 791.4 \\
 & 30 & -- & 29.221 & -- & 1,026.3 \\
\midrule
\shortstack[l]{LinkedIn\\Hiring\\Assistant} & 3 & 2.997 & 3.009 & 433.3 & 123.1 \\
 & 4 & 3.995 & 4.012 & 532.9 & 137.7 \\
 & 5 & 4.990 & 5.014 & 739.5 & 158.6 \\
 & 6 & 5.978 & 6.016 & 967.9 & 179.0 \\
 & 8 & 7.971 & 8.020 & 1,465.9 & 228.4 \\
 & 10 & 9.883 & 10.024 & 3,011.3 & 295.2 \\
 & 12 & 10.191 & 12.003 & 14,231.5 & 420.4 \\
 & 14 & -- & 13.959 & -- & 477.0 \\
 & 16 & -- & 15.935 & -- & 610.7 \\
 & 18 & -- & 18.013 & -- & 718.1 \\
 & 20 & -- & 19.996 & -- & 856.6 \\
 & 22 & -- & 21.849 & -- & 954.9 \\
 & 24 & -- & 23.543 & -- & 1,193.8 \\
 & 26 & -- & 25.297 & -- & 1,411.6 \\
 & 28 & -- & 27.085 & -- & 1,770.2 \\
\bottomrule
\end{tabular}
\end{table*}

The table explains the distinction between a latency-only result and a
passing joint-criterion point. In particular, the Qualification Retrieval
Platform's highest tested offered rate remains below the mean-p95 threshold but falls short of the completion
requirement. We report the highest measured passing point and do not
interpolate or extrapolate beyond the tested grid.

\section{Numerical Validation}
\label{app:agreement}

Validation separates exact constraints from floating-point execution.
Artifact construction reconstructs the accepted SID set and checks set
equality against its input. At serving time, every response is checked for
length, result count, duplicate SIDs, and membership in that same set.
These checks establish valid constrained execution, not identical ranking.

An independent oracle derives valid children directly from synthetic path
tuples and computes dense full-vocabulary log-softmax and top-\(k\).
The suite covers dense and sparse representations, varied fan-out,
beam widths below and above codebook size, and tied scores. The compact path
matches the oracle's selected tokens and log probabilities.

For the production-derived replay, we compare CLP and combined outputs on
matched prompts. Set overlap at cutoff \(k\) is
\[
  \frac{|S^{\mathrm{CLP}}_{i,k}\cap S^{\mathrm{combined}}_{i,k}|}{k}.
\]
\ifdefined\arxivpreprint
Table~\ref{tab:agreement}
\else
The set-agreement table
\fi
in the main paper reports means over 1,000 prompts per model.
The same application resolver is also applied to the first 50 prompts,
using its application cutoff. Those resolved-set results are agreement
measurements, not relevance evaluations.

Floating-point reduction and batch order can change near-tied scores even
when the real-arithmetic formulas are equivalent. We therefore do not infer
bitwise identity or unchanged retrieval quality from trie validity or
set agreement. Relevance labels are required to establish Recall, MRR,
or NDCG equivalence.

\section{Execution Configuration}
\label{app:reproducibility}

Both systems use one H200 per run, BF16 weights and KV cache, and tensor and
data parallelism of one. Table~\ref{tab:appendix-runtime} gives the
model-specific resource limits and warmup settings. Model weights,
tokenizer configuration, and the ordered prompt workload remain fixed
across the two implementations. Source revisions and implementation-specific
backend settings are recorded with the benchmark configuration.

\begin{table}[t]
\centering
\small
\caption{Replay configuration for the two serving systems.}
\label{tab:appendix-runtime}
\resizebox{\columnwidth}{!}{%
\begin{tabular}{lll}
\toprule
Setting & \shortstack[l]{Qualification\\Retrieval Platform} &
\shortstack[l]{LinkedIn\\Hiring Assistant} \\
\midrule
Static memory fraction & 0.75 & 0.55 \\
Maximum internal rows & 16,209 & 20,020 \\
Graph group limit & 9 & 10 \\
Prefix caching & Enabled & Disabled \\
Warmup requests & 20 & 50 \\
Measured requests per run & 1,000 & 1,000 \\
Independent runs per point & 5 & 5 \\
\bottomrule
\end{tabular}
}
\end{table}

The internal row budget is distinct from client concurrency: one expanded
beam request can occupy approximately \(B\) decode rows. Raising client
concurrency does not raise this budget. Each process initializes the engine
and its graph state before warmup; primary inference timing excludes both.
Mean intervals use the five run-level samples rather than treating individual
requests as independent repetitions. Ratio intervals independently resample
runs from each system.

The fixed-depth SID path excludes stop tokens from its codebooks. We verify
that condition against the model and tokenizer before enabling the
fixed-length selection path. Unsupported output semantics or attention
layouts use the fallbacks described in
\ifdefined\arxivpreprint
Section~\ref{sec:serving-integration}.
\else
the main paper's ``SGLang integration and fallbacks'' subsection.
\fi

\paragraph{Profiling.}
Nsight traces cover measured decoding after initialization and warmup.
They verify native constraint, attention, and selection execution and graph
replay. Short traces are diagnostic and do not replace the repeated
1,000-request capacity measurements.

\section{External Comparison Protocols}
\label{app:external}

The FlashRec comparison uses the complete Qualification Retrieval Platform
catalog, the same ordered input tokens, beam 200, and client concurrencies
4, 10, and 32. Both systems
use BF16 weights and KV cache, prefix caching, and CUDA graphs on one H200.
Each repetition measures 1,000 requests after 20 warmups. FlashRec contributes
two repetitions per point and the combined system contributes five.
Only the five-run means receive confidence intervals.

The score normalizers differ: the combined system uses the full vocabulary,
whereas FlashRec uses its supported active-codebook path. Consequently,
throughput and identifier validity can be compared at this configuration
without asserting score or ranked-output equivalence.

The GR-Inference comparison is restricted to a catalog compatible with its
native REC limits. We retain branches in deterministic lexicographic order,
limit fan-out to the supported bound, and retain terminal children without
performance-based selection. The same restricted catalog is encoded for
both systems and checked for set equality.

That comparison uses beam 1,000 and candidate-normalized scoring on both
systems. Each of two processes replays 1,000 requests after 50 warmups, with
Poisson arrivals at 62 QPS and no client-concurrency cap. The production
LinkedIn Hiring Assistant configuration exceeds the native beam-width and fan-out limits;
the compatible result is therefore not a throughput claim for the
unrestricted workload.
\fi

\end{document}